\pdfoutput=1

\documentclass[11pt]{article}

\usepackage{acl}

\usepackage{times}
\usepackage{latexsym}
\usepackage{graphicx}
\usepackage[T1]{fontenc}

\usepackage[utf8]{inputenc}
\usepackage{enumitem}

\usepackage{microtype}

\usepackage{inconsolata}

\usepackage{inconsolata}
\usepackage{titlesec}
\usepackage{hyperref}
\usepackage[labelfont=bf]{caption}
\title{Facts-and-Feelings: Capturing both Objectivity and Subjectivity in Table-to-Text Generation}

\author{Tathagata Dey \and Pushpak Bhattacharyya \\
  Department of Computer Science and Engineering, \\
  Indian Institute of Technology Bombay\\
  \texttt{tathagata@cse.iitb.ac.in} \\}

\begin{document}
\maketitle
\begin{abstract}
Table-to-text generation, a long-standing challenge in natural language generation, has remained unexplored through the lens of subjectivity. Subjectivity here encompasses the comprehension of information derived from the table that cannot be described solely by objective data. Given the absence of pre-existing datasets, we introduce the \textbf{Ta2TS dataset} with \textbf{3849 data instances}. We perform the task of fine-tuning sequence-to-sequence models on the linearized tables and prompting on popular large language models. We analyze the results from a quantitative and qualitative perspective to ensure the capture of subjectivity and factual consistency. The analysis shows the fine-tuned LMs can perform close to the prompted LLMs. Both the models can capture the tabular data, generating texts with \textbf{85.15\%} BERTScore and \textbf{26.28\%} Meteor score. To the best of our knowledge, we provide the first-of-its-kind dataset on tables with multiple genres and subjectivity included and present the first comprehensive analysis and comparison of different LLM performances on this task. 
\end{abstract}

\section{Introduction}

In the contemporary era of big data, humongous volumes of information are being generated and archived in numerous different formats. Among these, tables stand out as a very useful way of storing structured data. Using the potential of natural language to comprehend tabular data holds the promise of enhancing human efficiency across diverse applications. Text generation from tables has seen some contributions previously (See Section \ref{related}). These efforts have predominantly centred on either the generation of text from relatively simple tabular structures or the translation of numerical data into natural language, devoid of nuanced interpretation. In this research, we propose a subjective point of view to look at the table-to-text generation problem statement.


\begin{figure}[!ht]
    \centering
    \includegraphics[scale=0.68]{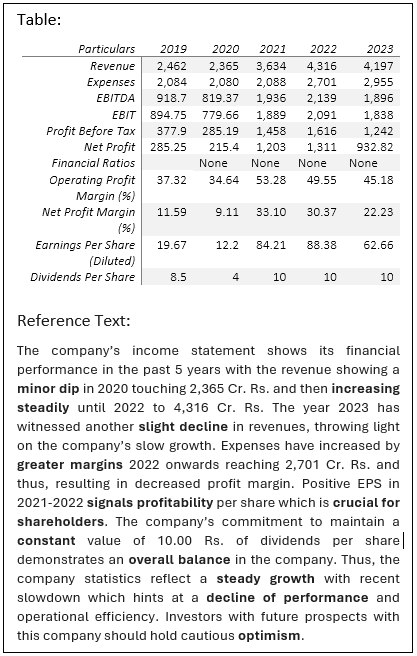}
    \setlength{\belowcaptionskip}{-16pt}
    \caption{\textbf{Generating text with subjectivity from a table}: The table contains the income statement report of a company over 5 years. The reference text below describes the tabular information where the bold phrases refer to the infused subjectivity. This subjectivity is the interpretation of the data from a human perspective. (Due to space constraints, shown in figure \ref{fig:subobj2})}
    \label{fig:subjobj}
\end{figure}

\subsection{Problem Statement}
\label{probst}
We introduce a novel problem statement with the aim of text generation from tabular data, devoted to infusion of subjectivity into the generated text. Subjectivity, in the context, refers to the nuanced interpretation of the data, beyond the realm of numeric or objective representation.

The table depicted in Figure \ref{fig:subjobj} presents a company's income statement table with 10 features such as Revenue, Expenses, EBIDTA, EBIT, Profit before taxes, Net profit, Financial ratios, Operating profit margin, Net profit margin, Earnings per share and Dividends per share respectively. Within the accompanying narrative of the table, several expressions are employed, which do not find direct representation in the objective data. Instead, these phrases refer to interpreting the underlying sentiment associated with the data.

The phrase \textbf{\emph{minor dip}} comes from the idea that having a downfall of 200 Cr. rs. is \textbf{small} compared to a growth of 2000 Cr. rs. over the last few years. Whereas, \textbf{\emph{signals profitability}} comes from the understanding that positive EPS refers to positive growth \footnote{The table example is taken from Groww official website}.

\subsection{Problem Formulation}

The input in this task is a table with multiple rows and columns (denoted as \(T\) here) and the output is a natural language text or sequence of tokens generated by the model (denoted as \(S\) here).
The problem of table-to-text generation takes a table \(T\) as input where \(T=\{t_{1,1}, ..,t_{i,j},..,t_{m,n}\}\) where \(t_{i,j}\) refers to the token of \(ith\) row and \(jth\) column with at most \(m\) rows and \(n\) columns. A pre-processed input document \(Y\) is produced from the table where \(Y\) can be written as, \(Y=y_1y_2..y_l\). Here \(y_i\) is the \(ith\) token in \(Y\) which has \(l\) tokens and \(l\ge mn\). In an auto-regressive approach text \(S=s_1s_2...s_k\) is generated from \(Y\) which can be defined as, \(s_i=argmax P(s_i|Y,s_1,s_2,...,s_{i-1}; \theta)\) where \(\theta\) is model parameters.

\subsection{Motivation}
Text generation from tables can be useful in various social applications where a mere description of the numerical values proves to be insufficient.


\begin{itemize}[noitemsep,nolistsep]
    \item News or Blog writings on a Tournament points table, Pricing table, or Voting results table.
    \item Reports on a Match summary, Business details, Sales detail of a company.
    \item Explaining Healthcare reports, Weather reports, or understanding Legal documents, etc.
\end{itemize}

In recent years Table-to-text generation has seen significant contributions (section \ref{related}). However, the domain of exploration has been very limited to sports and Wikipedia tables mostly. Along with the perspective of subjectivity, which can add a human touch to the generations, we explore different genres of tables, which is necessary to make comprehensive systems for this task.

The advancement of LLMs has contributed largely to text generation, but not so much in the case of tabular data. A study revealing the potential possibility of using LLMs for this task is also necessary in the current scenario.

\subsection{Our Contributions}
Our contributions are:
\begin{enumerate}[noitemsep,nolistsep]
    \item The formulation of a problem statement (Section \ref{probst}) of subjective text generation from tables, enriched with social significance. To the best of our knowledge, this is the first-of-its-kind effort.
    \item The \textbf{Ta2TS} (pronounced as Tattoos and expanded as Table-2-Text with Subjectivity and Objectivity) dataset for this task comprising \textbf{3849 table instances} sourced from sports, financial statements, and weather forecast domains (Section \ref{dataset}).
    \item Analysis and comparison of the \textbf{performance} of different \textbf{LLMs} on the dataset, enabling insight into LLMs' capability of maintaining factual consistency, coverage, and semantic similarity with reference for table-to-text generation task (Section \ref{llmeval}).
    \item Showing that regular LMs perform close to LLMs. This we do by developing a T5-based fine-tuned sequence-to-sequence learner which has the highest Bleu-4 and BERTScore on test Ta2TS dataset, \textbf{beating GPT3.5} and the other LLMs (Section \ref{metrictable}).
\end{enumerate}

\section{Related Works}
\label{related}
The table-to-text problem contains several different approaches based on the type of table. The \textbf{Wikibio-infobox} problem \cite{wikibiodata} involves generating textual descriptions from tables with just one column extracted from Wikipedia infoboxes. Versatile neural language models and encoders were introduced to tackle this challenge by \citet{wikibio1} and \citet{wikibio3}.

The \textbf{ToTTo} dataset \cite{tottodata} represents another prominent objective of text generation from a single highlighted row of the table. \citet{totto1} harnessed the T5 model to address this challenge. Subsequently,  sequence-to-sequence techniques and structure-aware frameworks were introduced by \citet{totto2} and \citet{totto3}. The current state-of-the-art on this dataset has been held by \citet{totto4}, who leveraged the T5 model with 3 billion parameters to achieve the highest scores. A very similarly aligned problem, \textbf{Wiki-table-to-text}, has also been worked on by \citet{wikitabletext1}.

\label{rotowire}
Among the previously discussed problem statements, none encompass the task of comprehending complicated tables with multiple rows and columns. The \textbf{Rotowire} dataset \cite{rotowiredata} is expressly designed to address this challenge. The concepts of macro-planning, content selection, and planning have been significant contributions to this field by \citet{roto1}.  \citet{roto5} achieved state-of-the-art results by introducing a Record encoder, a Reasoning module, and a Decoder equipped with Dual attention. Additionally, alternative approaches have also been explored by \citet{roto4}, \cite{roto6} and \citet{roto3}.  It is worth noting that the RotoWire tables exhibit a notable degree of similarity to the tables of our interest. The flattened tables derived from data instances within RotoWire can contain up to 17,000 tokens. Most of the current language model architectures do not support such a huge context length. Thus, using pre-trained models for this task is very challenging. While training a model from the ground up may address this challenge, there are concerns regarding its comprehensive understanding of the language. \citet{llm-tqa} explored the \textbf{performances of LLMs} in table question-answering tasks with comparative analysis. Similar studies in text generation from tables could be useful in the future.

\section{Dataset}
\label{dataset}
We contribute to building a novel dataset for this problem which is named \textbf{Ta2TS Dataset} (Table-to-Text with Subjectivity), pronounced as \emph{Tattoos}.

\subsection{Data Collection and Filtration}
\label{datacoll}
The dataset encompasses three types of tables, such as financial statement tables, weather forecast tables, and sports tables. We collected the tables using various web-based platforms that are entitled to showcase different data tables. Five sites were selected as sources of the tables such as \textit{Groww}, \textit{Indian Meteorological Dept.}, \textit{ESPN Cric Info}, \textit{IPL}, and \textit{Goal} (see \ref{datacollection} for sources). The tables are scraped using Python Programming language. A total of 4200 tables were collected from the sources by web scraping. The collected data has been filtered manually. Tables with fewer rows and columns than a threshold are removed. Also, data instances having very low cell values are discarded. Such as, some of the collected tables had zero or NaN values. After filtration, the number of instances in the dataset stands at 3849. The complete dataset statistics are in Table \ref{tab:datadis}. The dataset contains arguably an unbiased distribution between different genres, shown in Figure \ref{fig:datastat1}.

\subsection{Annotation Details}
\label{annotation}
We employ three annotators for this task who are proficient in English. One of the annotators is a graduate of Computer science and engineering while the others are PhD students and postgraduates in English literature respectively. They were provided with detailed annotation guidelines with rules, objectives, and multiple examples. 
Each instance in the set was annotated only once. The distribution among the annotators was 1150, 1300, and 1399 instances. An example annotation is shown in figure \ref{fig:gen_ex} (See \ref{appannot}). 100 samples were annotated by all three annotators which are used to compute the \textbf{Inter-Annotator Agreement} (IAA). The annotators were asked to mark the other annotators' annotations \footnote{Annotator A is asked to mark annotators B and C's annotations for the given features.} based on the correctness of factual information and appropriate use of subjectivity out of 10 (1 being the lowest score and 10 being the highest). Also, pairwise BERT-based semantic similarity has been captured to emphasize the similarity in the annotation tasks (scaled down to 10). BERTScore ensures the contextual meaning of the sentences. We present the harmonic mean of these two metrics as the Inter-Annotator Agreement. We have also evaluated pairwise Bleu-2 score for all the annotators which is mentioned in the appendix (section \ref{appannot}).

\begin{table}[!ht]
    \centering
    \begin{tabular}{c|c c c c}
          & \textbf{A} & \textbf{B} & \textbf{C} & \textbf{D}\\
          \hline
         \textbf{A} & & 8.74 & 8.64 & 8.43\\
         \textbf{B} & 8.56 & & 8.99 & 8.42\\
         \textbf{C} & 8.69 & 8.84 & & 8.67\\
         \textbf{D} & 8.33 & 8.69 & 8.99 & \\
    \end{tabular}
    \caption{\textbf{Inter-Annotator Agreement}: We compute the IAA using two metrics. Pairwise scoring of each annotation based on factual consistency and capture of subjectivity and calculation of pairwise average BERTScore. The table represents the harmonic mean of these two metrics.}
    \label{tab:iaa}
\end{table}

\begin{table*}[ht]
    \centering
    \begin{tabular}{r c c c}
    \hline
        Genre & No. of Instances & Avg. no. of rows & Avg. no. of columns \\
        \hline
        \textbf{\large Finance} & & & \\
        Income statement & 500 & 11 & 5\\
        Balance Sheet & 499 & 11 & 5\\
        Cash flow & 499 & 11 & 5\\
        \hline
        \textbf{\large Weather forecast} & & & \\
        District-wise tables & 1031 & 5 & 10\\
        \hline
        \textbf{\large Sports} & & & \\
        Tournament Points Table & 409 & 9.4 & 7.2\\
        Series Form Table & 941 & 5 & 6 \\
        \hline
        \textbf{Total} & \textbf{3849} &   & \\
        \hline
    \end{tabular}
    \caption{\textbf{Genres in the Ta2TS Dataset}: The Ta2TS dataset comprises three primary genres, namely, financial tables, weather forecast tables and sports tables. In this table number of tables from each domain and different types are shown.}
    \label{tab:datadis}
\end{table*}

\section{Experimental Setup}

We performed experiments on different versions of the T5 model \cite{t5}. T5 is a sequence-to-sequence model trained on C4 data with prefixes. We understand that the metadata associated with each table (company, team or district name, forecast range etc.) can be an essential component of the text generation. We fine-tuned the T5 models with different sets of prefixes as a part of our ablation study to understand the importance of table context for the model.

\subsection{Linearization of Tables}
\label{linearization}
Transformer-based sequence-to-sequence models are capable of understanding linearized information with long context. A closer examination of the reference texts reveals that the tabular content contains two distinct types of information, namely, row-wise and column-wise comparison. For example, the income-statement table depicts phrases that express the relative growth of the company in time which requires row-wise comparison and relation between revenue-EBIDTA or other features which requires column-wise comparison. Hence, the application of row-major flattening along with a column separator token is required. In this linearization, each cell is separated by a \textbf{<sep>} token and rows start and end with \textbf{<row>} and \textbf{</row>} tokens respectively. An example of linearization is shown in section \ref{applin}.

\begin{figure*}[ht]
    \centering
    \includegraphics[scale=0.4]{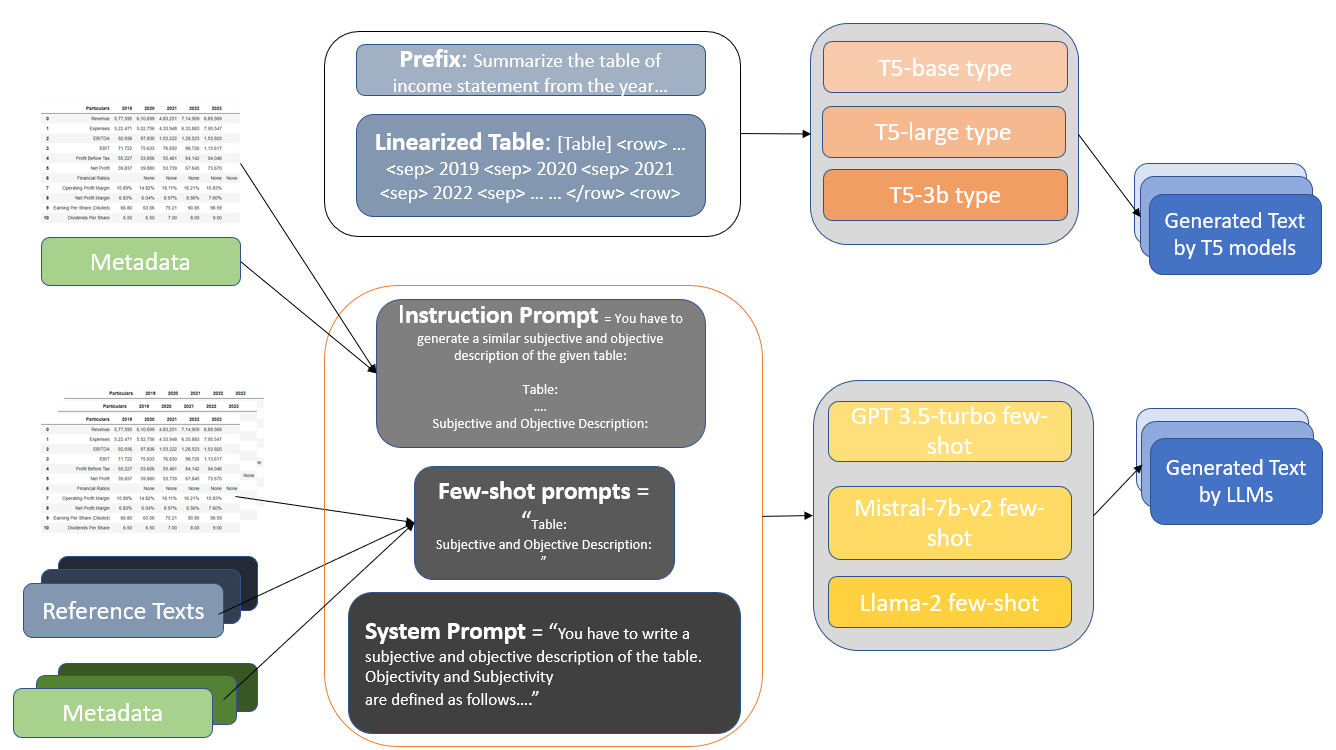}
    \caption{\textbf{Architecture}: The diagram shows two parallel methods of experimentation we adopted. In the top experiment, the tables are linearized (section \ref{linearization}). Metadata is the genre of the table, company or district name, time and additional details. Different types of prefixes are given based on the context (section \ref{prefixtypes}). Different T5 models are used with different prefix types. In the bottom experimentation, some popular LLMs are prompted. A system prompt is generated by formally defining the task. Multiple examples are given in few-shot methods along with the metadata and reference text. Finally generated texts by all of these methods are compared and analysed.}
    \label{fig:archi}
\end{figure*}

\subsection{Sequence-to-sequence Model}

We use the T5 \cite{t5} models to perform the task. T5 is a transformer-based sequence-to-sequence learner. Multiple versions of the T5 model such as T5-base, T5-large, and T5-3b are used for the experiments. We used three different setups with respect to prefixes, 
\label{prefixtypes}
\begin{enumerate}[noitemsep]
    \item Without any table context or prefix.
    \item With the prefix \emph{"Summarize the table: "}
    \item With contextual prefix such as \emph{"Summarize the table about Kolkata district of West Bengal state, during the time range 01-Jan-2024 to 06-Jan-2024: "}
\end{enumerate}

The overall architecture with linearization, prefix and model usage is shown in figure \ref{fig:archi}.

\subsection{Training Details}
The T5 models are fine-tuned on different settings. The dataset has been divided into an 80-10-10 train-test-validation split. Using five-fold cross-validation we trained the T5-base, T5-large and T5-3b models on 30, 24 and 20 epochs respectively. The number of epochs has been chosen by minute observation of the change in loss and metrics over the epochs. We used a learning rate of \(2e-5\) with a batch size of \(4\). On average, the whole fine-tuning for the base model took around 6 hours, for large around 10 hours and for 3b for 17 hours. Adafactor optimizer has been used with a learning rate of \(1e-3\) and with a decay rate of \(0.8\).

\subsection{Prompting on LLM}
\label{llmeval}
Along with T5 results, we perform prompting experiments on various state-of-the-art Large language models. Comparing and analysing the results we try to establish a baseline on this dataset. With the experiments on LLM, we fundamentally try to answer three different questions.
\begin{enumerate}[noitemsep]
    \item How well do LLMs perform in understanding human perception behind the tabular information?
    \item Can LLMs maintain factual consistency through the generation tasks?
    \item What is the future of table-to-text generation through generic LLMs?
\end{enumerate}
Generic LLM here refers to the models which are generally used for a wide variety of tasks and not specifically designed for a particular task, such as ChatGPT, Mistral etc. To answer the questions raised above, we have prompted on four Language models, namely, \textbf{GPT-3.5-turbo} \cite{gpt}, \textbf{Mistral-7b} \cite{mistral}, and \textbf{Llama-2} \cite{llama}. The experiments include 0-shot, 1-shot and few-shot (up to 7 shots). 


\label{prompttypes}
We experimented with different prompts with changing definitions of subjectivity and instruction. We divide the prompt into three sections. \emph{System prompt} is the part where we define the task and provide the necessary definitions to the LLM (see the figure \ref{fig:archi}). In the \emph{few-shot prompt} section, we add the example instances with notation of table and description. In the last \emph{instruction prompt} section, the input table with instructions to write description text is given. Our experiments focused on the system prompt section mostly. The tables are given in input with separator tokens and proper indentation. An example of a table and prompt is given in Figure \ref{fig:prompt_ex} (section \ref{prompting}).

\section{Results}

We measured the generated texts of the fine-tuned models and LLMs through automatic and human evaluation methods. The automatic metrics are explained below.

\subsection{Quantitative Analysis}
The three different prefixes are denoted by the following notations.

\label{prefixes}
\begin{enumerate} [noitemsep]
    \item \textbf{w/o-pf}: Without any table context or prefix.
    \item \textbf{pf-w/o-ct}:With the prefix \emph{"Summarize the table: "}
    \item \textbf{pf-ct}: With contextual prefix such as \emph{"Summarize the table about Kolkata district of West Bengal state, during the time range 01-Jan-2024 to 06-Jan-2024: "}
\end{enumerate}

\label{metrictable}

\begin{table*}[!ht]
    \centering
    \begin{tabular}{ r c c c c c }
        \hline
        \textbf{Model} & \textbf{Parent-Fscore} & \textbf{Bleu-4} & \textbf{Meteor} & \textbf{Rouge-L} & \textbf{BertScore}\\
        \hline
        T5-base, w/o-pf & 0.44 & 2.89 & 22.17 & 21.11 & 80.43 \\
        T5-base, pf-w/o-ct & 0.44 & 3.22 &  24.60 &  22.97 & 84.14\\
        T5-base, pf-ct & 0.45 & \textbf{3.25} &  24.98 &  23.45 & \textbf{85.15} \\
        \hline
        T5-large, w/o-pf  & 0.47 & 2.81 &  23.54 & 21.98 & 81.43\\
        T5-large, pf-w/o-ct  & 0.47 & 2.94 &  23.65 & 22.31 & 82.22 \\
        T5-large, pf-ct  & 0.48& \underline{3.20} & 24.11 & \textbf{23.67} & 84.37 \\
        \hline
        T5-3b, w/o-pf  & 0.41 & 2.57 & 20.11 & 21.39 & 81.58 \\
        T5-3b, pf-w/o-ct  & 0.42 & 2.63 & 21.77 & 22.64 & 82.00\\
        T5-3b, pf-ct  & 0.44 & 2.67 & 22.03 & 22.78 & 82.28\\
        \hline
        GPT3.5, 0-shot & 0.51 & 2.64 & 25.3 &  21.9 & 81.70 \\
        GPT3.5, 1-shot & \underline{0.52} & 3.08 & \textbf{26.28} & 23.02 & 83.87 \\
        GPT3.5, 3-shot & \textbf{0.54} & 2.98 & \underline{25.97} & \underline{23.57} & \underline{84.78} \\
        GPT3.5, 7-shot & 0.51 & 2.48 & 25.29 & 23.24 & 84.42 \\
        \hline
        Mistral-7b, 0-shot & 0.42 & 2.21 & 18.09 & 13.65 & 30.01 \\
        Mistral-7b, 1-shot & 0.42 & 2.48 & 18.76 & 18.11 & 40.83 \\
        Mistral-7b, 3-shot & 0.43 & 2.69 & 19.26 & 18.76 & 29.64\\
        Mistral-7b, 7-shot & 0.42 & 2.08 & 20.17 & 13.92 & 29.87 \\
        \hline
        Llama-2, 0-shot & 0.45 & 2.17 & 17.44 & 19.32 & 56.71\\
        Llama-2, 1-shot & 0.46 & 2.31 & 18.13 & 19.45 & 59.00\\
        Llama-2, 5-shot & 0.46 & 2.84 & 19.65 & 19.88 & 63.14\\
        \hline
    \end{tabular}
    \setlength{\belowcaptionskip}{-11pt}
    \caption{\textbf{Metrics Table}: We computed Parent-Fscore \cite{parentscore}, Bleu-4 score \cite{bleu}, Meteor score \cite{meteor}, Rouge-L score \cite{rouge} and BERTScore \cite{bertscore} to analyse the generations. The highest scores in each metric are highlighted in bold and the second highest is highlighted with underline. We can see that the results of our T5 model and comparable with GPT3.5 and they together outperformed the rest. Parent score is shown out of 1 and the rest of the metrics are shown out of 100.}
    \label{tab:quanti}
\end{table*}

\subsubsection{Evaluation Metrics}

\textbf{Parent-FScore} Parent-Fscore is F1 score that assesses the overall generation in the domain of table-to-text \cite{parentscore}. It uses entailment-based precision and recall to analyse the factual consistency in the generated texts. 

\textbf{Bleu-4 Score} It is an n-gram precision-based metric \cite{bleu}. B-4 calculates precision up to 4 grams and represents the lexical match between reference and machine-generated text.

\textbf{Meteor Score} It is an alignment-based metric which uses lexical matching, stemming-based matching, synonyms etc \cite{meteor}. It also penalizes unaligned words. Meteor represents a good characteristic for judging content and fluency.

\textbf{Rouge-L Score} It is a recall-based metric which measures the longest common sub-sequence between the machine-generated text and reference text \cite{rouge}. It emphasizes on how much content is covered from the reference text.

\textbf{BERTScore} This metric is based on contextual embeddings of the BERT model \cite{bertscore}. BERTScore represents the similarity in the reference text and generated text.

The results are showcased in table \ref{tab:quanti}. We can see that T5 and GPT3.5 outperformed all the other models. The T5-base (pf-ct) achieved the highest Bleu-4 score of \textbf{3.25} and the highest BERTScore of \textbf{85.15}. On the other hand, T5-large (pf-ct) achieved the highest Rouge-L score of \textbf{23.67} and the second-highest Bleu-4 of \textbf{3.20}. BERTScore signifies the similarity of generated text in the embedding space and B-4 represents the lexical content overlap.

As a part of the ablation study, we fine-tuned the T5 models with different prefix settings. As we can see from table \ref{tab:quanti}, the \emph{pf-ct} setting (section \ref{prefixes}) had the highest performances in both T5-base and T5-large. Hence, we conclude that adding the context in prefixes helps generate more relevant texts to the ground truth. For T5-3b, we estimate that it is overfitting on the training data. The generated texts from the 3b model lacked coverage significantly and sometimes hallucinating too.

\begin{figure*}[t]
    \centering
    \includegraphics[scale=0.65]{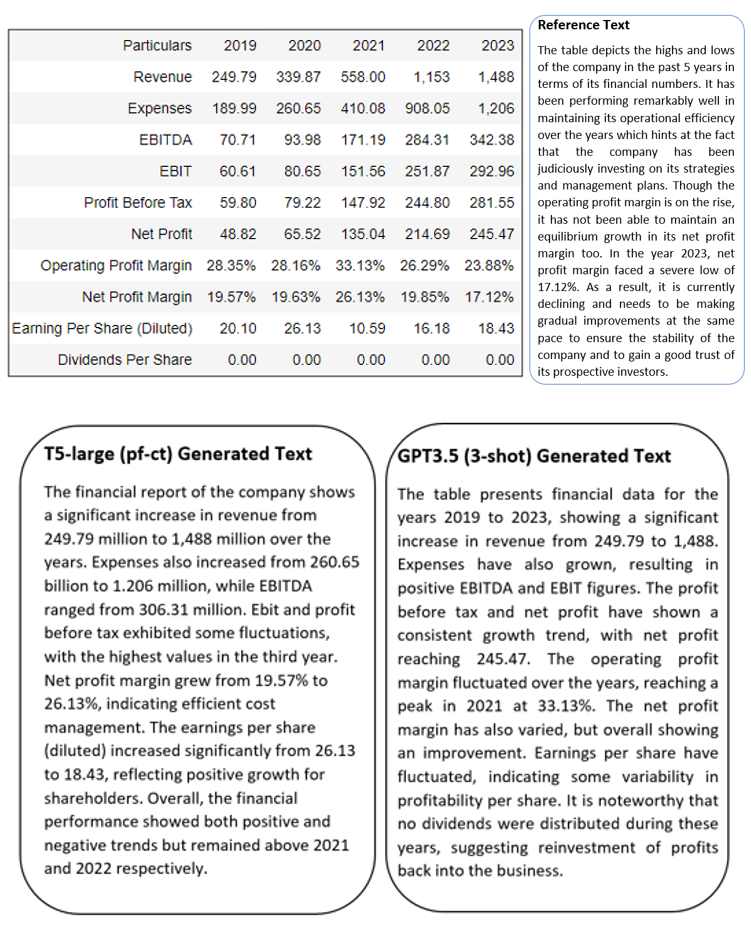}
    \caption{\textbf{Example of texts generated by T5-large (pf-ct) and GPT 3.5 (3-shot)}: In the figure the given table and reference text (ground truth) are taken from the Ta2TS dataset and at the bottom the generated texts are shown.}
    \label{fig:gen_ex}
\end{figure*}

GPT3.5 significantly outperformed all other large language models. With a 3-shot setting, it achieved the highest Parent-fscore and the second-highest scores in Meteor, Rouge-L and BERTScore. The generations covered all necessary information from the tables, had fluency, and coherence and provided significant similarity with the ground truth too. Mistral-7b and LLama-2 performed poorly on all the settings with similar prompts. For simple text generation, we understand that 3-shot prompting outperforms 7-shot. The shorter context helps the model understand the task and semantics of ground truth soundly. The B-4 scores are significantly low yet the generated texts represent a good summary of the table. We explain this by the property of synonyms. The T5 models and LLMs understand the language phenomena soundly. Hence, it can use different words or expressions to express similar meanings. Hence, the lexical overlap can be low. However, a similar meaning refers to close enough values of context vectors in BERT embedding space, which in turn is reflected in high BERTScores. An example of generated text is shown in Figure \ref{fig:gen_ex}.

\subsection{Qualitative Analysis}

\label{qualitativeanal}

Qualitative Evaluation has been performed over 100 randomly chosen samples. The evaluators are asked to mark the generations out of 10 (1 being the lowest and 10 being the highest), based on four parameters, namely, coherence, coverage, accuracy, and subjectivity capture. 

\textbf{Coherence} is defined as the logical flow of information in the text, where each sentence or phrase follows the previously generated phrase. 

\textbf{Coverage} is defined as ensuring the inclusion of all the necessary information from the table. Necessary, although depends on different perspectives, we have specifically defined to the evaluators. There can be information in the table which is co-related to some other information (can be inferred from other information), we call it dependent information. For instance, when we say that a team has played 16 matches and won 10 while losing 4, it is obvious that 2 matches were tied.
The evaluators are asked to ignore the dependent information and judge based on only the independent ones. 

\textbf{Accuracy} is defined as the factual correctness of the objective information present in the table. It is noteworthy that the accuracy metric will not penalize coverage, hence it will judge only the information in the generated text.

\textbf{Subjectivity capture} is defined as the correctness of subjective phrases generated in the text and the inclusion of subjective phrases for every objective information. 



The human evaluation scores on GPT3.5, 3-shot are shown in Table \ref{tab:gpteva}. The evaluators analysed each instance on the four features described above. To summarise scores, we report the harmonic mean of all the features as it represents a more refined differentiation than the arithmetic mean. The scores showcase that in all of the given features, the generations reflect soundness and quality. The relatively lower coherence is caused due to some unattended co-reference resolution in the text, which reduces the flow of reading to some extent. The high subjectivity capture score reflects on use of correct adjectives at the right places throughout the text.

\begin{table}[ht]
    \centering
    \begin{tabular}{c c c c c}
        \hline
         & \textbf{Coh.} & \textbf{Cov.} & \textbf{Acc.} & \textbf{Sub. Cap.} \\
        \hline
        \textbf{A} & 8.7 & 9.0 & 8.8 & 9.1\\
        \textbf{B} & 8.4 & 8.8 & 8.9 & 9.0\\
        \textbf{C} & 8.9 & 9.0 & 9.1 & 9.1\\
        \hline
        \textbf{H-Mean} & 8.66 & 8.93 & 8.93 & 9.06\\
        \hline
    \end{tabular}
    \setlength{\belowcaptionskip}{-10pt}
    \caption{\textbf{Human evaluation on GPT3.5 Generated Instances}: the average score of all 200 generated samples over the features of Coherence (Coh.), Coverage (Cov.), Accuracy (Acc.) and Subjectivity Capture (Sub. Cap.). We report the summary of scores by harmonic mean of all the evaluators.}
    \label{tab:gpteva}
\end{table}

The human evaluation scores of T5-large (pf-ct) generated instances, as shown in Table \ref{tab:t5eva}, indicate a lower quality compared to texts generated by GPT. Specifically, the coverage score for T5-generated texts is significantly lower than that of GPT-generated texts. Upon closer examination, it becomes apparent that many of the generated samples leave some objective facts unattended, which could have enhanced their impact on readers. Despite this, T5-generated texts achieve relatively higher scores in Accuracy and Subjectivity capture, indicating a factually consistent presentation of objective data and an appropriate emotional interpretation of the objective information context. However, some issues with co-reference resolution persist in these texts like GPT, which results in lower coherence scores.

\begin{table}[ht]
    \centering
    \begin{tabular}{c c c c c}
        \hline
        & \textbf{Coh.} & \textbf{Cov.} & \textbf{Acc.} & \textbf{Sub. Cap.} \\
        \hline
        \textbf{A} & 8.3 & 8.2 & 8.7 & 8.8\\
        \textbf{B} & 8.1 & 8.0 & 8.5 & 9.0\\
        \textbf{C} & 8.4 & 8.3 & 8.8 & 9.0\\
        \hline
        \textbf{H-Mean} & 8.26 & 8.16 & 8.66 & 8.93\\
        \hline
    \end{tabular}
    \setlength{\belowcaptionskip}{-10pt}
    \caption{\textbf{Human evaluation on T5-large-pf-ct Generated Instances}: The average score of all 200 generated samples over the features of Coherence (Coh.), Coverage (Cov.), Accuracy (Acc.) and Subjectivity Capture (Sub. Cap.). We report the summary of scores by harmonic mean of all the evaluators.}
    \label{tab:t5eva}
\end{table}

\section{Conclusion and Future Work}

In this paper, we present a novel perspective on the domain of table-to-text generation by introducing the element of subjectivity, an unexplored dimension of this field for several different table genres. To facilitate our goal, we curate the \textbf{Ta2TS} Dataset with tables and respective subjective-objective ground truth texts. We conduct fine-tuning and prompting experiments to assess the effectiveness of various models in capturing subjectivity and factual objectivity. The evaluations indicate that the models exhibit a robust understanding of word knowledge, with subjective phrases appropriately reflected in the generated text. Our fine-tuned T5 models perform comparably with the GPT3.5-turbo (few-shot prompted) model. Also, we provide a to-the-best-of-our-knowledge first comprehensive analysis of LLM performances in the table-to-text generation task.
We proposed three questions regarding the use of LLMs in table-to-text generation (section \ref{llmeval}) which we try to answer here. The detailed quantitative and qualitative evaluations reveal that the generated texts are sound in terms of most of the evaluation parameters. LLMs do understand the structure of tabular data and the relation between different cells. The texts remain consistent factually throughout, indicating reliability in the objective information. Also, it has higher coverage than the fine-tuned models. Hence, we conclude the effectiveness of LLMs in table-to-text generation.

\section{Limitations}
Our dataset contains tables from three genres. Subjectivity in the same type of table can be influenced by similar semantics, which may not be the case for a more diverse dataset. Hence, building a robust system to handle all kinds of tables will be more feasible with an even more diverse or large dataset. Moreover, we developed a table flattening technique to fit tables in sequence-to-sequence models which are trained on normal text. For prospects, we propose ideas to fuse pre-trained table encoders with decoders and fine-tune our dataset to get even better metric scores.

\section{Ethics Statement}
All of our collected data were present in open-source mediums and did not contain personal, restricted, or illegal data. None of the generated texts or annotated texts were intended to promote or derogate any team or entity.

\bibliography{custom}

\appendix

\section{Appendix}
\label{sec:appendix}

\subsection{Examples of Subjective Text Generation from Tables}
\label{appexm}

\begin{figure}[!ht]
    \centering
    \includegraphics[scale=0.42]{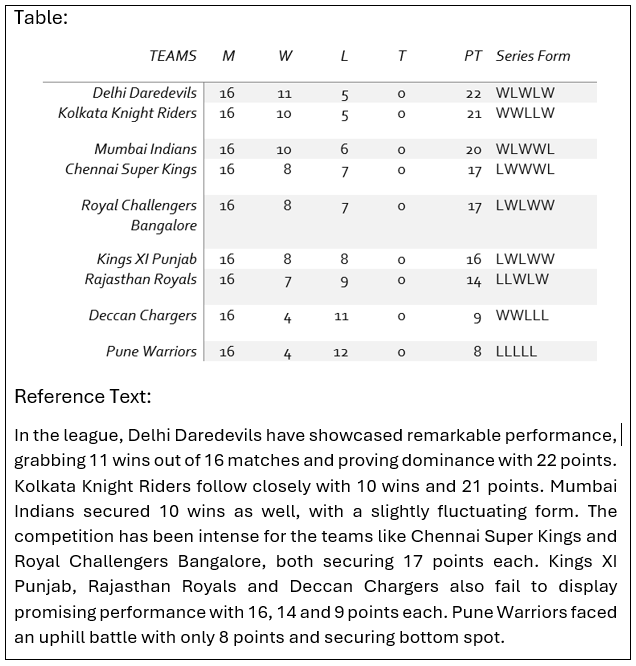}
    \caption{\textbf{Generating text with subjectivity}: The table shows different teams and their relative performances in a tournament. The table is described by the reference text, enriched with subjectivity. It is an example instance table from the sports domain of the \textbf{Ta2TS dataset}. In the table the abbreviations denote the following; M:Matches; W:Wins; L:Losses; T:Ties; PT:Points.}
    \label{fig:subobj2}
\end{figure}

\begin{figure*}[t]
    \centering
    \includegraphics[scale=0.55]{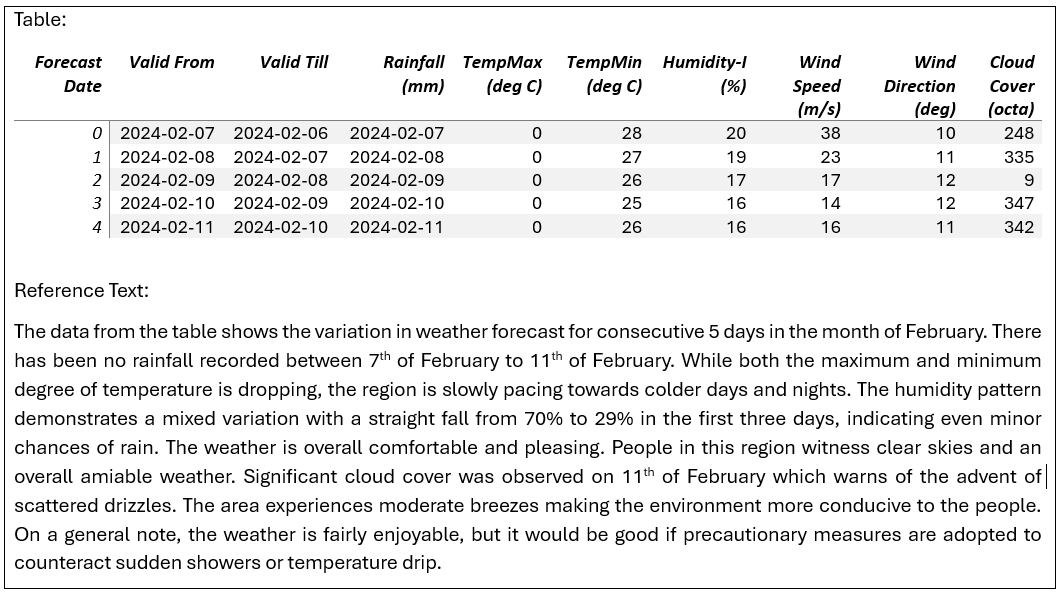}
    \caption{Generating text with subjectivity: The table contains data on the weather forecast in the district of Kolkata from 6th February 2024 to 11th February 2024. The subjective reference text describes the table. This example data instance is taken from the \textbf{Ta2TS dataset}.}
    \label{fig:weathertable}
\end{figure*}

Figure \ref{fig:subobj2} and figure \ref{fig:weathertable} showcase some more examples of data instances from the \textbf{Ta2TS dataset}. These tables are collected from Sports and Weather forecast domains respectively.

\subsection{Data Collection}
\label{datacollection}
The tables of Ta2TS Dataset have been scraped from 5 sites namely, Groww \footnote{\url{https://groww.in/}}, Indian Meteorological Dept. \footnote{\url{https://mausam.imd.gov.in/imd_latest/contents/agromet/advisory/indiadistrictforecast.php}}, ESPN Cric Info \footnote{\url{https://www.espncricinfo.com/}}, IPL \footnote{\url{https://www.iplt20.com/}} and Goal  \footnote{\url{https://www.goal.com/en-in}} which are mentioned in the section \ref{datacoll}.

\subsection{\textbf{Ta2TS} Dataset Statistics}
\label{appstat1}
\begin{figure}
    \centering
    \includegraphics[scale=0.8]{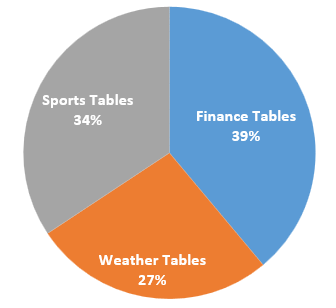}
    \caption{\textbf{Distribution of Genres in Ta2TS Dataset}: The chart depicts the distribution of different genres in the dataset. 34\% of the tables are from the sports genre, 39\% are from the finance genre and 27\% are from the weather genre.}
    \label{fig:datastat1}
\end{figure}

The dataset contains three different kinds of tables such as financial-statement tables, weather forecast tables, and sports tables. The financial statements tables are of different companies that are listed on the National Stock Exchange, India. There are three kinds of financial tables for each company, namely its income statement, balance sheet, and cash flow tables. On the other hand, the weather forecast data is collected from the official site of Govt. of India. These tables present data for each district in each state. Statistics of the dataset have been shown in table \ref{tab:datadis} and figure \ref{fig:datastat1}. As mentioned in section \ref{datacoll}, a total of 3849 tables are collected with an average of nearly 9 rows and nearly 7 features. 

\subsection{Dataset Annotation}
\label{appannot}
The data instances have been manually annotated which is mentioned in section \ref{annotation}. The annotators were given a detailed instruction set to follow and were paid a total of 1.20\$ per data instance for annotation. The annotation task was conducted through regular monitoring over two months, following a fixed set of instructions. We asked the annotators to write at most 150 words for each instance. We try to keep the length of the reference text invariable for a variable number of rows and columns.

We define the annotation task in three steps: (1) Information extraction from the table, (2) Subjectivity infusion, and (3) Text generation. In step 1, the annotators are expected to extract the objective information in the table and understand its dynamics, changes, and relations. Furthermore, they should understand the emotions lying with the variables in the data and infuse subjectivity into the information. In the last step, they should write a text combined with the previous semantics which describes the table. The annotation details and guidelines are described in section \ref{annotation}. The pairwise Bleu-2 scores are shown in table \ref{tab:iaab2}.

\begin{table}[!ht]
    \centering
    \begin{tabular}{c|c c c c}
          & \textbf{A} & \textbf{B} & \textbf{C} & \textbf{D}\\
          \hline
         \textbf{A} & & 27.13 & 32.49 & 29.33\\
         \textbf{B} & & & 23.53 & 34.18\\
         \textbf{C} & & & & 28.58\\
         \textbf{D} & & & & \\
    \end{tabular}
    \caption{\textbf{Bleu-2 for Inter-Annotator Agreement}: Pairwise Bleu-2 scores are calculated for annotators.}
    \label{tab:iaab2}
\end{table}

\subsection{Table Linearization}
\label{applin}
Linearization of tables is crucial when it is given as input in seq-to-seq models. We used a tagged concatenation of delimiter-separated row-major ordering to generate a flattened sequence of tables. The row-major orders are initiated with tags \textbf{<rows>} and end with \textbf{</rows>}. Whereas, each column in row-major order is separated by \textbf{<sep>} token. The intuition behind the linearization is described in section \ref{linearization}. 

\begin{figure}[!ht]
    \centering
    \includegraphics[scale=0.48]{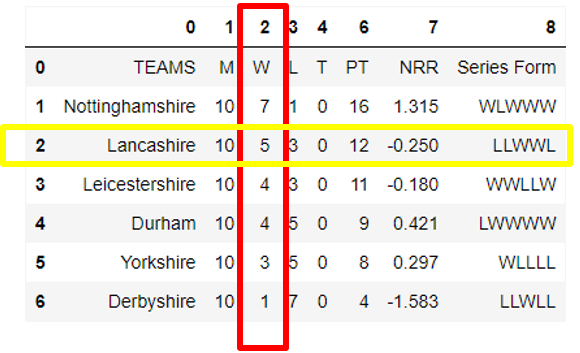}
    \setlength{\belowcaptionskip}{-12pt}
    \caption{Table flattening: extracting information from row-major and column-major ordering}
    \label{fig:flattening}
\end{figure}

A row-major flattening of the table in figure \ref{fig:flattening} can be written as, \emph{[TABLE] <row> Teams <sep> Matches <sep> Wins <sep> Losses <sep> ... <sep> Series Form </row> <row> Nottinghamshire <sep> ... <SEP> ... LLWLL </rows>}.

\subsection{Prompting on LLMs}
\label{prompting}

\begin{figure*}
    \centering
    \includegraphics[scale=0.7]{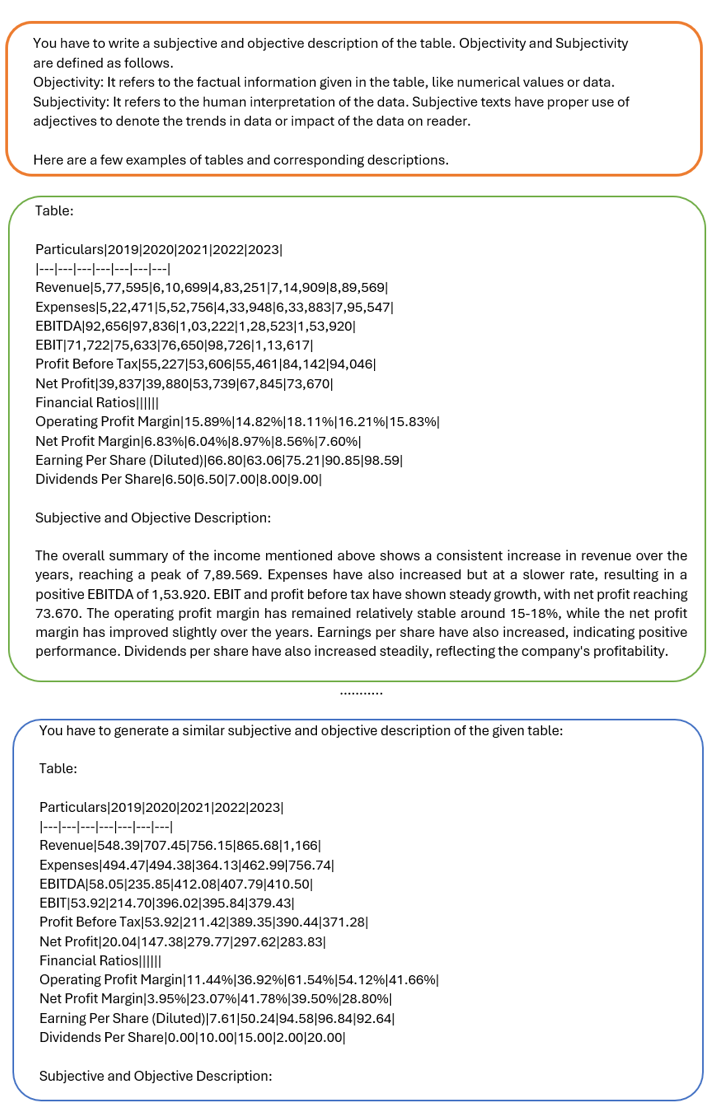}
    \caption{\textbf{Prompt Example}: An example prompt to the LLMs is given in the figure. The three sections represent system prompt, few-shot prompt and instruction prompt explained in section \ref{prompttypes}. The few-shot section is repeated several times depending on the shot numbers.}
    \label{fig:prompt_ex}
\end{figure*}

Large language models are prompted to generate objective and subjective text from the given tables. We use three different sections in the whole prompt. The first section is \textbf{system prompt} which refers to the initial description of the task along with necessary definitions. Followed by \textbf{few-shot prompt}, where multiple examples of tables and description texts are given. This section can be repeated several times depending on the shot numbers. In the last section, i.e., \textbf{instruction prompt}, we add the input table along with a little instruction about what has to be performed. An example of prompt has been shown in the figure \ref{fig:prompt_ex}.

\end{document}